# Image Registration for Stability Testing of MEMS

Nargess Memarsadeghi[*a], Jacqueline Le Moigne[a], Peter N. Blake[a],
Peter A. Morey[b], Wayne B. Landsman[c], Victor J. Chambers[a], Samuel H. Moseley[a].

[a]NASA Goddard Space Flight Center, Greenbelt, MD, 20771.
[b]Ball Aerospace and Technologies Corp., Lanham, MD, 20706.
[c]Adnet Systems Inc., Rockville, MD 20852.

## ABSTRACT

Image registration, or alignment of two or more images covering the same scenes or objects, is of great interest in many disciplines such as remote sensing, medical imaging, astronomy, and computer vision. In this paper, we introduce a new application of image registration algorithms. We demonstrate how through a wavelet based image registration algorithm, engineers can evaluate stability of Micro-Electro-Mechanical Systems (MEMS). In particular, we applied image registration algorithms to assess alignment stability of the *MicroShutters Subsystem* (MSS) of the *Near Infrared Spectrograph* (NIRSpec) instrument of the James Webb Space Telescope (JWST). This work introduces a new methodology for evaluating stability of MEMS devices to engineers as well as a new application of image registration algorithms to computer scientists.

## 1. INTRODUCTION

The James Webb Space Telescope (JWST) is a large aperture space telescope designed to provide imaging and spectroscopy from 1.0μ to 29μ [1,2] - a spectral bandpass which requires the instrument to be cooled to 32-35 Kelvin. The Goddard Space Flight Center (GSFC) is the lead center for the JWST program and manages the project for NASA [2,3]. NIRSpec is an instrument that will allow scientists to capture the spectra of more than 100 objects at once [3]. Since the objects NIRSpec will be looking at are so distant and faint, the instrument needs a way to block out the light of near bright objects as needed. This is the function of a new technology called Microshutters. Microshutters are tiny aperture cells that measure $100\mu \times 200\mu$ and are arranged in a grid of about $1.5 \times 1.5$ inches, that contains over 62,000 of them [3, 4]. Evaluating and verifying the stability and alignment requirements of optical MEMS devices [5], such as Microshutters, via automatic image registration algorithms is a novel approach. This approach was extensively exercised by our team for the MSS.

Various stability requirements of the MSS had to be verified and confirmed under various conditions. These requirements include the stability of the subsystem with respect to changes in horizontal and vertical positions when the subsystem went through *cryogenic cycles* in GSFC's Cryogenics Research and Integration Facility (CRIF). In other words, MSS locations should be measured and tracked when the system cools down from the room temperature, 295K-235K, to temperatures as low as 20K-40K, and as it heats up again.

An astronomical quality high-definition camera was utilized to image the MSS, which had a cruciform structure in center and four grids of Microshutters on the adjacent corners. Round bright targets were laser-etched into the black coating of the cruciform. The grids had to serve as their own targets, since nothing could be attached to them. The camera would take 10-30 successive high resolution images, similar to Figures 1(a-b) without the labels and marked regions. Images were of size 4096 × 4086 pixels, where each pixel represents 3μ in the spatial domain. We refer to the four Microshutter grid regions as *Quads* 1-4. Bright filled disks of varying size represent the laser etched targets on different parts of the cruciform plate, serving as features for image registration algorithms. We call these *Crux* regions. Various Quad and Crux regions are labeled in Figure 1(a).

---

[*] Corresponding author: Nargess Memarsadeghi, NASA GSFC, Mail Stop 587, Greenbelt, MD, 20771. Email: Nargess.Memarsadeghi@nasa.gov. Tel: 301-286-2938.

Different image registration algorithms were used to compare movements of different regions of these images against corresponding regions in a baseline image. We were interested in measuring the relative Quad-to-Crux movements before, during, and after the subsystem went through heat cycles. The relative movements of the MSS Quad regions with respect to the Crux regions were required to be within 0.4μ of the predicted thermal dimensional changes. We discuss two of the image registration algorithms that were used to measure and verify the stability requirements of the MSS. One method was based on the wavelet decomposition of images, and the other based on their cross-correlation values. The rest of this paper is organized as follows. In Section 2 we give an overview of image registration and mention the registration algorithms that were used for our application. We describe our experiments in Section 3. Results are presented in Section 4. Section 5 concludes the paper.

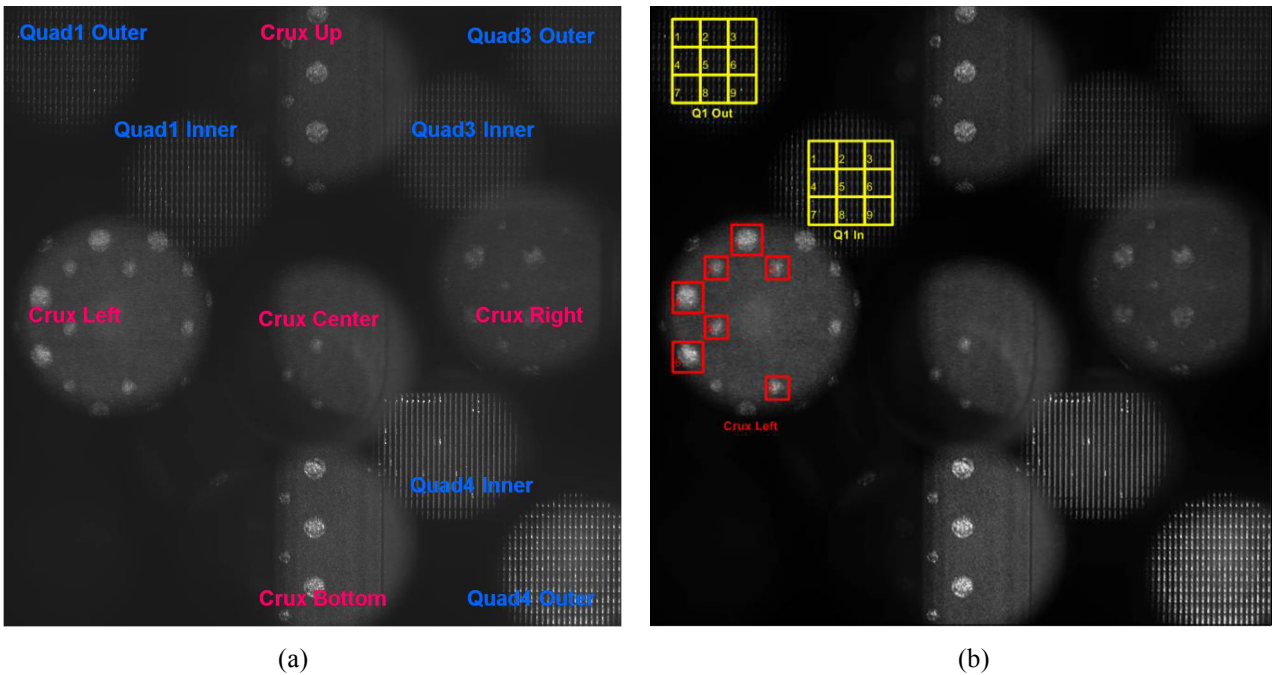

(a) (b)

Figure 1- Sample images with (a) Quad and Crux regions labeled, and (b) sample regions used for the XREG algorithm marked.

## 2. IMAGE REGISTRATION ALGORITHMS

An image registration algorithm calculates a geometric transformation of a certain type that best aligns a given pair of images [6]. Some registration algorithms are *manual,* meaning that they rely on user's selection and matching of a set of *control points* or *regions of interest.* These algorithms then use this information to solve for a desired transformation, which then will be applied to the input image to be aligned to the reference image. *Automatic* registration algorithms, however, are of great interest since they enable *unsupervised* batch registration of many pairs of images without requiring prior knowledge of the images. We used an automatic image registration algorithm extensively and verified its results with those obtained from a manual registration algorithm for stability analysis of the MSS alignment after going through extreme temperature changes.

**Wavelet-Based Registration (TRU):** The automatic registration algorithm, used in this paper, relies on wavelet decomposition. Similarly to a Fourier transform, wavelet transforms provide space-frequency representations of a *2D* signal, which can be inverted for later reconstruction. However, conversely to a Fourier or even a windowed Fourier transform, wavelets also provide a better spatial localization as well as a better division of the space-frequency plane. For practical applications for which the signal is not infinite in extent, it provides better space details especially for high frequencies (e.g., around region edges in images).

In a wavelet representation, the original signal is filtered by the translations and the dilations of a basic function, called the "mother wavelet". Equation 1 shows the general continuous form of a wavelet transform of an image $I$,

$$Wav(I)(a,b) = \frac{1}{\sqrt{a}} \int_{-\infty}^{\infty} \int_{-\infty}^{\infty} I(u,v) . W(\frac{u-b_1}{a}, \frac{v-b_2}{a}) d_u d_v , \qquad (1)$$

where $W$ represents the "mother wavelet", $b=(b_1, b_2)$ is the translation factor and $a$ is the dilation factor. All the dilations and translations of the mother wavelet form an orthonormal basis in which the function image is uniquely represented and therefore the transformation can be inverted to produce the original image from the unique representation. A similar equation is given in the discrete domain with:

$$Wav(I)(a,b) = \sum_m \sum_n a^{-m/2} I(x,y) . W(a^{-m}x - nb_1, a^{-m}y - nb_2). \qquad (2)$$

Typical values for $a$ and $b$ are $a=2$ and $b=1$.

In 1989, Mallat demonstrated the connection between filter banks and wavelet basis functions [9]. His results became very important for the application of wavelet theory to image processing applications. Specifically, for a 2D image, this work shows that wavelet analysis can be implemented by filtering the original image by a high-pass and a low-pass filter, iteratively in a multiresolution fashion; then, depending on the type of wavelets, whether the filters are 1D or 2D, a decision is made as to whether images should be decimated after each level of decomposition or not, etc. Generally, features provided through wavelet decomposition can be described as being of two different types: the low-pass features which provide a compressed version of the original data as well as some texture information, and the high-pass features which provide detailed information very similar to edge features. The process is iterated at each level by decomposing the low-pass image (or "compressed" version of the original image), thus building a hierarchy of lower and lower resolution images. Since at each iteration the size of the compressed image decreases, by leaving the size of the filter unchanged, the area being filtered in the compressed image corresponds to a larger and larger portion of the original image.

For various applications, different types of wavelets have been proposed [7]. In this work, we used Spline wavelets [8]. This type of wavelet was chosen for its *invariance* properties. That is, the wavelet transform of a shifted or rotated image would be the same as the shifted or rotated wavelet transform of the original image. This is a very important property when using wavelets for an image registration application.

The automatic registration algorithm, used for the MSS stability analysis, was initially developed by Thevenaz, Ruttiman and Unser [10] for registering 3D medical images. It is based on an $L2$-based least-squares multiresolution optimization, using a modified version of the Marquardt-Levenberg (ML) algorithm [11]. The ML algorithm represents a hybrid optimization approach between a pure gradient-descent method and the Gauss-Newton method, more powerful but less robust. In the remainder of this paper, we refer to this registration algorithm as "TRU". The original TRU algorithm is implemented using a coarse-to-fine Spline pyramid. This algorithm was later modified and included in an image registration toolbox for registering large 2D remotely-sensed satellite imagery at NASA GSFC [6]. For the work presented in this paper, we used this modified version with low-pass features provided by the Spline wavelets [8]. The inputs to the algorithm are the 2D gray scale input and reference images with their dimensions. These dimensions need to be a multiple of 2$l$, where $l$ is the number of decomposition levels used by the registration algorithm. We used subset images of size 512 × 512 for registering various quad and crux regions similar to those displayed in Figure 2. The registration is performed in a multiresolution fashion starting from the lowest resolution images for which an affine transformation is computed between the reference low-pass features and the input low-pass features. Then, the affine transformation is iteratively refined at each resolution level (from low to high) until the highest spatial resolution is reached. At each level, a Marquart-Levenberg optimization is used to find the best transformation.

**Correlation-Based Registration (XREG):** The XREG algorithm uses cross-correlation matching and was developed in IDL based on the XREGISTER routine available within the IRAF[†] package [11]. This program was developed for applications in optics and astronomy. In particular, the program does the best job in matching points (e.g. point sources such as stars). This algorithm requires the user's input in manually identifying and matching some features in the reference and input images. Therefore, one can hardly apply this technique over thousands of images, as we could with TRU approach. However, this approach was used in many cases to verify both stability of the MSS system and the performance of the TRU algorithm for our application. Movements of different areas were calculated using the XREG approach by first identifying and matching nine features between the reference and input images manually and putting a box around them. Figure 1(b) demonstrates an example of boxes that were selected for registering the Crux Left regions in middle left; nine boxes were selected for registering each of the inner and outer Quad regions in upper left section of the image. After selecting and matching the features, the registration was already performed manually within 1-3 pixels accuracy. Then, the discrete convolution of the reference and input image regions over the window of interest, typically a 5×5 box, was calculated for each region. In order to find the location of the peak value within the sub-pixel accuracy, a *2D* parabolic function was fitted to the convolution results. Therefore, for each crux and quad area, nine horizontal shift *(Tx)* and vertical shift *(Ty)* values, peak locations of each feature, were calculated. The XREG routine reports the median of these values as the final result. Results of this algorithm also served as reference or proof of the accuracy of the automatic TRU algorithm for this new application.

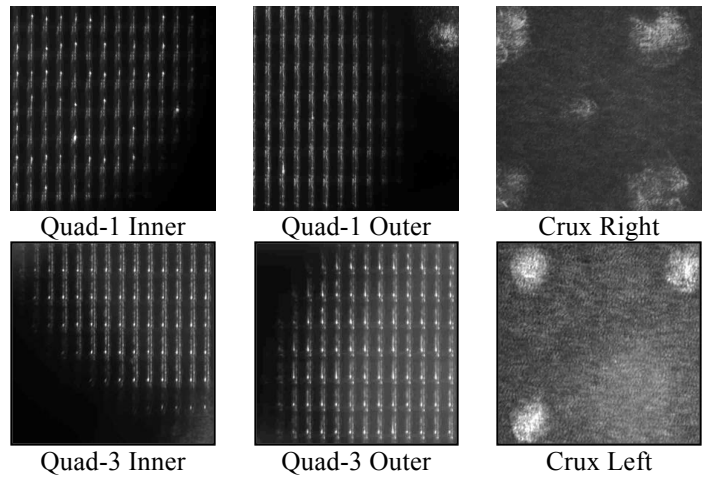

| Quad-1 Inner | Quad-1 Outer | Crux Right |
| Quad-3 Inner | Quad-3 Outer | Crux Left |

Figure 2- Sample regions extracted from images for TRU.

## 3. EXPERIMENTS

The system was tested over several *heat cycles*. At the beginning of each cycle, 30 images, each of size 4096 × 4086 similar to Figures 1(a-b), were collected. These images had the resolution of 3μ per pixel. Then the system would heat up to 260K (shutter reset condition). The system would later go through a cool-down to reach 32K-35K (cryogenic operating condition), at which point 30 post cool-down images were obtained. This cycle was repeated six times during January 16-23, 2009. The system's stability was measured via image analysis approaches during these cycles. We considered the first image obtained at 33K on January 17 (post cool-down on the 16th and pre warm-up on the 17th) as our reference image. We compared all other images obtained at the cryogenic condition, 330 images from January 17-23, to this reference image. We used the TRU and the XREG image registration algorithms for our analysis.

---

[†] IRAF (Image Reduction and Analysis Facility) is distributed by the National Optical Astronomy Observatories, which are operated by the Association of Universities for Research in Astronomy, Inc., under cooperative agreement with the National Science Foundation.

For the TRU approach, 11 regions each of size 512 × 512, similar to those in Figure 2, were extracted from all 330 images. Each of these regions was registered to their corresponding region extracted from the reference image. That is, the TRU algorithm ran 330 × 11 = 3630 times only for the data presented in this paper. The TRU algorithm reported the transformation values of the input images with respect to the reference image, horizontal and vertical movements *(Tx* and *Ty* respectively) and the rotation angle *(tetha),* for each case.

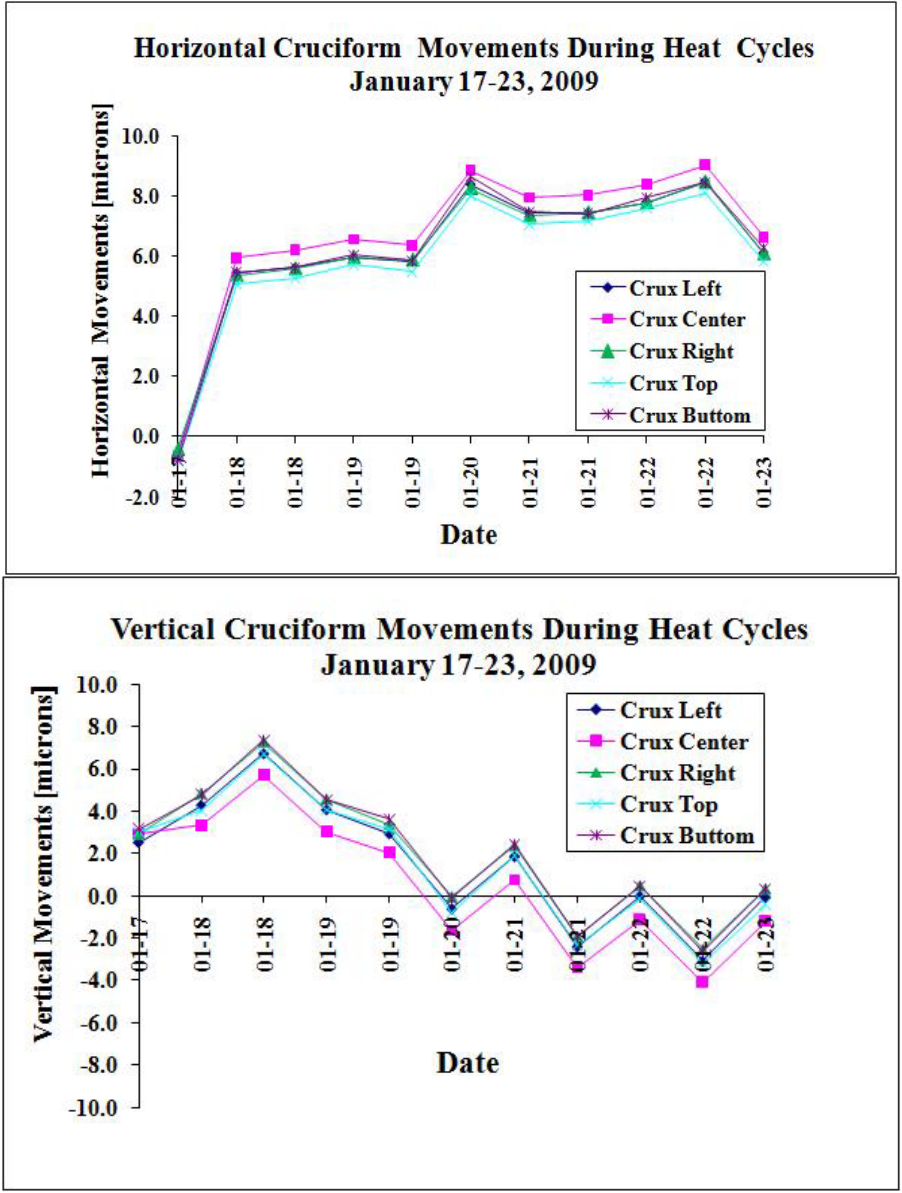

Figure 3- Average daily Crux movements from TRU algorithm.

For the XREG approach, only the Crux Left, Inner and Outer Quad 1 regions of the 330 raw images were analyzed (990 runs of the XREG). This is while for each region, 9 boxes containing features were identified and matched, similar to marked boxes in Figure 1(b). The algorithm reported the median of the selected nine features' horizontal (*Tx*) and vertical (*Ty*) movements with respect to the reference image.

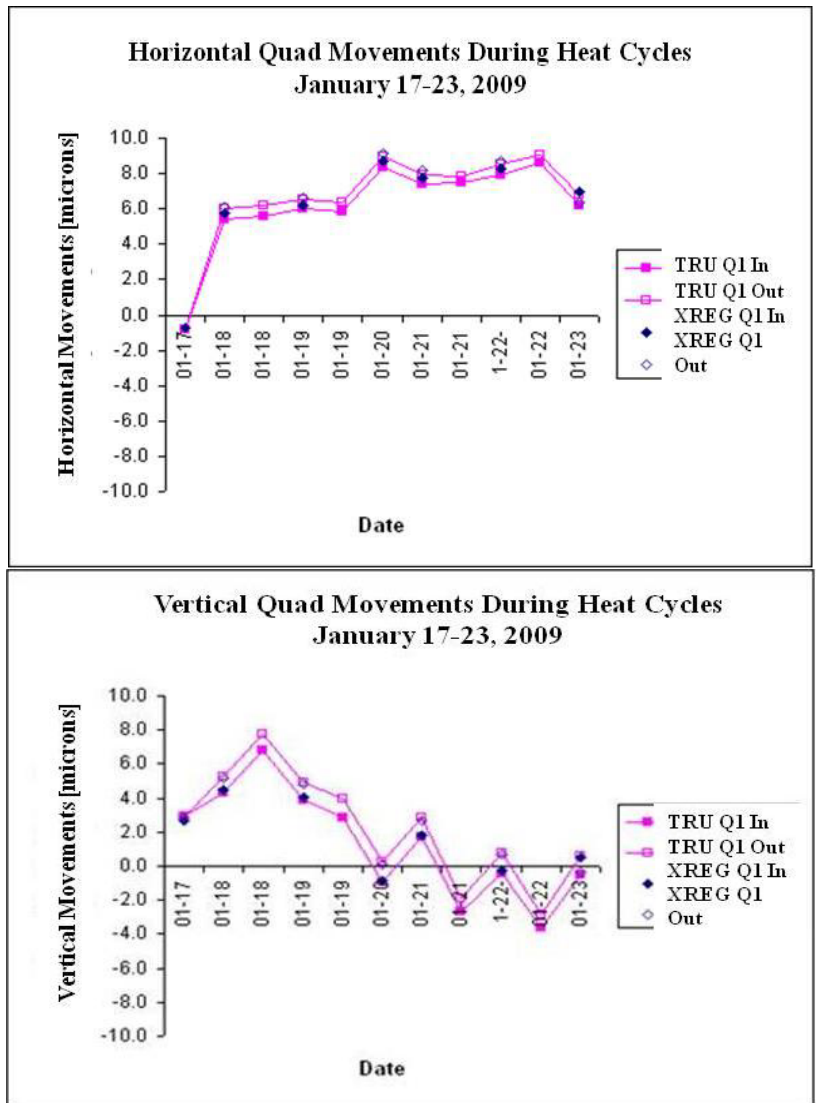
Figure 4- Comparison of TRU and XREG results

## 4. RESULTS

We calculated the movements of all 13 regions (similar to those in Figure 2) extracted from 330 MSS images of January 17-23 with respect to the reference image using TRU algorithm. Figure 3 demonstrates these horizontal and vertical movements for Crux regions. These values are averaged over 30 images of each day. As expected, the calculated shifts for all Crux regions are very close to each other. Similar measurements and results were obtained via TRU for Quad regions. Next, we compared the above mentioned results from the TRU algorithm with those obtained from the XREG approach. We had comparable results from both methods when registering the Inner Quad 1, Outer Quad 1, and Crux Left regions. Figure 4 demonstrates that while the movements of the inner and outer Quad 1 regions are in almost all cases within less than a micron of each other, results obtained from TRU and XREG approaches are almost identical and overlapping. The two algorithms performed similarly for the Crux Left regions. After verifying the correctness and accuracy of the automatic TRU method via the supervised XREG algorithm, the MSS stability was calculated through the TRU results. First, the relative movements of average daily quad shifts with respect to the overall daily crux movements were calculated. As Figures 3 and 4 demonstrate the crux and quad regions' movements differed less than a pixel from each other. In fact, the maximum relative quad-crux shift we calculated was 2.83µ, a little less than one pixel. Then, each quad's movement was calculated by averaging over the movements of its corresponding inner and outer regions. After more detailed analysis, which requires a separate report, the MSS team demonstrated the stability of the MSS alignment within 0.3µ, or a tenth of a pixel's resolution.

## 5. CONCLUSION

The stability of the MSS was measured within 0.3µ. This measurement became possible via high resolution imaging, image analysis, and the TRU algorithm that allowed fast batch registration of hundreds of images at sub-pixel accuracy. We verified the results of the TRU wavelet-based registration algorithm with those obtained from a correlation based technique, which first required manual registration of features by users. Our successful experience suggests that our automatic image registration algorithm can be used for stability verification of MEMS devices.

## 6. ACKNOWLEDGMENTS

We thank Barbara Zukowski and Dr. Alexander Kutyrev of the MSS team for their help and support in acquiring raw images in the CRIF lab. We also thank Dr. Ilya Zavorin for answering our many questions on the TRU algorithm.

## REFERENCES

[1] J. P. Gardner, J. C. Mather, M. Clampin, R. Doyon, M. A. Greenhouse, H. B. Hammel, J. B. Hutchings, P. Jakobsen, S. J. Lilly, K. S. Long, J. I. Lunine, M. J. McCaughrean, M. Mountain, J. Nella, G. H. Rieke, M. J. Rieke, H. Rix, E. P. Smith, G. Sonneborn, M. Stiavelli, H. S. Stockman, R. A. Windhorst, and G. S. Wright, "The James Webb Space Telescope," *Space Science Reviews*, vol. 123, no. 4, pp. 485–606, April 2006.
[2] M. Clampin, "The James Webb Space Telescope (JWST)," *Advances in Space Research*, vol. 41, no. 12, pp. 1983–1991, 2008.
[3] A. S. Kutyrev, N. Collins, J. Chambers, S. H. Moseley, and D. Rapchun, "Microshutter arrays: High contrast programmable field masks for JWST NIRSpec." in *Space Telescopes and Instrumentation 2008: Optical, Infrared, and Millimeter.*, ser. SPIE, J. M. Oschmann Jr., M.W. M. de Graauw, and H. A. MacEwen, Eds., vol. 7010, pp. 70103D–1:10, 2008.
[4] R. F. Silverberg, R. Arendt, D. E. Franz, G. Kletetschka, A. Kutyrev, M. J. Li, S. H. Moseley, D. A. Rapchun, S. Snodgrass, D. W. Sohl, and L. Sparr, and the Microshutter Team, "A microshutter-based field selector for JWST's multi-object near infrared spectrograph," in *Infrared Spaceborne Remote Sensing and Instrumentation XV*, ser. SPIE, M. Strojnik-Scholl, Ed., vol. 6678, pp. 66 780Q–1:8, 2007.
[5] T. R. Hsu, *MEMS and Microsystems: Design, Manufacture, and Nanoscale Engineering,* 2nd ed. Hoboken, New Jersey: John Wiley & Sons, Inc., March 2008.
[6] I. Zavorin and J. Le Moigne, "Use of multiresolution wavelet feature pyramids for automatic registration of multisensor imagery," *IEEE Transactions on Image Processing*, vol. 14, no. 6, pp. 770-782, June 2005 .
[7] I. Daubechies, *Ten lectures on wavelets*, ser. CMBS-NSF regional conference series in applied mathematics. Philadelphia, PA: Society for Industrial and Applied Mathematics (SIAM), 1992.
[8] M. Unser, A. Aldroubi, and M. Eden, "The L2-polynomial spline pyramid," *IEEE Transactions on Pattern Analysis and Machine Intelligence*, vol. 15, no. 4, pp. 364–379, April 1993.
[9] S. G. Mallat, "A theory for multiresolution signal decomposition: the wavelet representation," *IEEE Transactions on Pattern Analysis and Machine Intelligence (PAMI)*, vol. 11, no. 7, pp. 674–693, July 1989.
[10] P. Thevenaz, U. E. Ruttiman, and M. Unser, "A pyramid approach to subpixel registration based on intensity," *IEEE Transactions on Image Processing*, vol. 7, no. 1, pp. 27–41, January 1998.
[11] D. W. Marquardt, "An algorithm for least-squares estimation of nonlinear parameters," *Journal of the SIAM Journal on Applied Mathematics*, vol. 11, no. 2, pp. 431–441, 1963.
[12] D. Tody, "IRAF in the Nineties," in *Astronomical Data Analysis Software and Systems II*, ser. ASP Conference Series, R. J. Hanisch, R. J. V. Brissenden, and J. Barnes, Eds., vol. 52, pp. 173–183, 1993.